%% file: main.tex
\documentclass[sigconf]{acmart}

\input{meta/cmd.tex}
\AtBeginDocument{%
  \providecommand\BibTeX{{%
    \normalfont B\kern-0.5em{\scshape i\kern-0.25em b}\kern-0.8em\TeX}}}

\setcopyright{acmlicensed}
\copyrightyear{2018}
\acmYear{2018}
\acmDOI{XXXXXXX.XXXXXXX}

\acmConference[Conference acronym 'XX]{Make sure to enter the correct
  conference title from your rights confirmation emai}{June 03--05,
  2018}{Woodstock, NY}
\acmISBN{978-1-4503-XXXX-X/18/06}

\usepackage{soul}
\definecolor{dypink}{HTML}{ec008c}

\begin{document}

\input{meta/title.tex}
\input{meta/authors.tex}

\input{content/0_abstract_keywords.tex}


\received{20 February 2007}
\received[revised]{12 March 2009}
\received[accepted]{5 June 2009}

\maketitle

\input{content/1_intro.tex}
\input{content/2_related_work.tex}
\input{content/3_preliminaries.tex}
\input{content/4_model.tex}
\input{content/5_exp.tex}
\input{content/6_discussion.tex}
\input{content/7_conclusion.tex}

\input{meta/acks}

\bibliographystyle{ACM-Reference-Format}
\bibliography{meta/ref}

\appendix

\onecolumn

\input{content/a1_glossary_table.tex}
\input{content/a2_data.tex}
\input{content/a3_exp_results.tex}

\end{document}

%% file: meta/cmd.tex
\graphicspath{{figures/}{./}}

\usepackage{times}
\usepackage{helvet}
\usepackage{courier}
\usepackage{natbib}
\usepackage{caption}
\usepackage{newfloat}
\usepackage{listings}
\usepackage{enumitem}
\usepackage[OT1,T1]{fontenc}
\usepackage{bm}
\usepackage{algorithm}
\usepackage[noend]{algorithmic}
\usepackage{amsmath}
\usepackage{bbding}
\usepackage{tabu}
\usepackage{booktabs}
\usepackage{colortbl}
\usepackage{multirow}
\usepackage{multicol}
\usepackage{color}
\usepackage{ulem}

\newcommand{\parasum}[1]{\textbf{\textit{#1}}}
\newcommand{\circled}[1]{\textcircled{\raisebox{-0.9pt}{#1}}}
\newcommand{\algcmt}[1]{\STATE // \textit{#1}}

\newcommand{\dsr}[3]{& #1 & #2 & #3 \\}
\newcommand{\gtr}[3]{#1 & #3 & #2 \\ }
\newcommand{\vta}[0]{\begin{tabular}[c]{l}Ranged\end{tabular}}
\newcommand{\vtb}[0]{\begin{tabular}[c]{l}Unlimited\end{tabular}}
\newcommand{\vtc}[0]{\begin{tabular}[c]{l}Boolean\end{tabular}}
\newcommand{\vtd}[0]{\begin{tabular}[c]{l}Categorical\end{tabular}}
\newcommand{\ablt}[2]{\midrule\multirow{#2}{*}{#1}}
\newcommand{\best}[1]{\textcolor{orange}{\textbf{#1}}}
\newcommand{\cmpr}[7]{\texttt{#1} & #2 & $\pm$ & #3 & #4 & $\pm$ & #5 & #6 & $\pm$ & #7 \\}
\newcommand{\cmptsr}[7]{\texttt{#1} & #4 & $\pm$ & #5 & #6 & $\pm$ & #7 \\}

\newcommand*\ablr[8]{%
 \gdef\ca{#1}%
 \gdef\cb{#2}%
 \gdef\cc{#3}%
 \gdef\cd{#4}%
 \gdef\ce{#5}%
 \gdef\cf{#6}%
 \gdef\cg{#7}%
 \gdef\ch{#8}%
 \ablrcont
}
\newcommand\ablrcont[6]{& \cb & \ca & \cc & \cd & \ce & \cf & \cg & \ch & #1 &$\pm$& #2 & #3 &$\pm$& #4 & #5\% & #6\\}

\newcommand\projurl{https://anonymous.4open.science/r/CATP-0710}

%% file: meta/title.tex
\title{CATP: Context-Aware Trajectory Prediction with\\Competition Symbiosis}

%% file: meta/authors.tex

\author{Jiang Wu}
\affiliation{
  \institution{Zhejiang University}
  \city{Hangzhou}
  \country{China}
}
\email{wujiang5521@zju.edu.cn}

\author{Dongyu Liu}
\affiliation{
  \institution{UC Davis}
  \city{Davis}
  \state{CA}
  \country{USA}
}
\email{dyuliu@ucdavis.edu}

\author{Yuchen Lin}
\affiliation{
  \institution{Zhejiang University}
  \city{Hangzhou}
  \country{China}
}
\email{yuchenlin@zju.edu.cn}

\author{Yingcai Wu}
\authornote{Yingcai Wu is the corresponding author.}
\affiliation{
  \institution{Zhejiang University}
  \city{Hangzhou}
  \country{China}
}
\email{ycwu@zju.edu.cn}

\renewcommand{\shortauthors}{Wu et al.}

%% file: content/0_abstract_keywords.tex
\begin{abstract}
Contextual information is vital for accurate trajectory prediction.
For instance, the intricate flying behavior of migratory birds hinges on their analysis of environmental cues such as wind direction and air pressure.
However, the diverse and dynamic nature of contextual information renders it an arduous task for AI models to comprehend its impact on trajectories and consequently predict them accurately.
% To unleash the full potential of contextual information, we propose CATP, a Context-Aware Trajectory Prediction model.
% CATP comprises a manager model, several worker models, and a tailored training mechanism inspired by competition symbiosis in nature.
% During training, workers compete against each other for training data and learn unique moving patterns.
To address this issue, we propose a ``manager-worker'' framework to unleash the full potential of contextual information and construct CATP model, an implementation of the framework for Context-Aware Trajectory Prediction.
The framework comprises a manager model, several worker models, and a tailored training mechanism inspired by competition symbiosis in nature.
Taking CATP as an example, each worker needs to compete against others for training data and develop an advantage in predicting specific moving patterns.
The manager learns the workers' performance in different contexts and selects the best one in the given context to predict trajectories, enabling CATP as a whole to operate in a symbiotic manner.
% We conducted two comparative experiments, an ablation study, and a case study to comprehensively evaluate CATP model, which proved that CATP could outperform SOTA models and be generalized to different context-aware tasks.
We conducted two comparative experiments and an ablation study to quantitatively evaluate the proposed framework and CATP model.
The results showed that CATP could outperform SOTA models, and the framework could be generalized to different context-aware tasks.
% A case study showed how the manager selected workers and the predicted trajectories of CATP, demonstrating the interpretability of the proposed framework.
\end{abstract}


\ccsdesc[500]{Computing methodologies~Machine learning algorithms}
\ccsdesc[300]{Computing methodologies~Neural networks}

\keywords{context-aware trajectory prediction, training algorithm, ensemble learning}

%% file: content/1_intro.tex
\section{Introduction}

Trajectory prediction is widely applied in many scenarios, such as traffic forecast \cite{fang2021mdtp,lu2022Vehicle}, pedestrians' movements prediction \cite{bartoli2018context}, and sports player position prediction \cite{lindstrom2020predicting}.
In most scenarios, contextual information can significantly affect the trajectories.
For example, pedestrians will give way to each other and avoid obstacles \cite{bartoli2018context}.
Thus, many trajectory prediction models integrate contextual information to improve prediction accuracy.

However, as big data science evolves, contextual information has become diverse and dynamic, bringing new challenges.
\textit{Diversity}:
The trajectory is affected by a combination of different types of contextual factors.
For example, birds' migratory trajectories are affected by wind direction, temperature, air pressure, etc.
Multiple contextual factors collectively and interactively influence the trajectories.
\textit{Dynamic}:
The contextual factors are ever-changing, leading to varied moving strategies and resulting trajectories. 
For example, birds may change their migratory routes as the wind directions change, since birds prefer to fly with a tailwind \cite{liechti1994effects}.
Video game players may change their trajectory when their opponents move to different places, to implement their counter competition plan.

Given such rich contextual information, AI models usually struggle to maximize its use to improve prediction accuracy, especially when we lack a large amount of labeled data recording how the contextual information affects trajectories.
Models based on single components tailored for encoding contextual information (SC) \cite{lu2022Vehicle,yu2021dynamic,lindstrom2020predicting} can accurately learn the impacts of contexts on trajectories.
However, such models suffer from increasingly complex contextual characteristics \cite{dong2020survey}.
Models based on ensemble learning (EL) \cite{li2022ensemble,jiang2017spatial,zhang2019prediction} train multiple base learners on manually split sub-datasets and combine their results by weights, increasing prediction accuracy in diverse contexts.
However, the performance of EL-based models depends heavily on how to split the dataset and assign weights reasonably, requiring a high level of domain expertise.
Models based on reinforcement learning (RL) design rewards based on contexts and predict the trajectories that can gain the highest rewards.
However, designing a reward function aligned with the predicted objects' individual preferences for moving is challenging.

% our work
To tackle these challenges, we propose \textbf{CATP}, a \underline{C}ontext-\underline{A}ware \underline{T}rajectory \underline{P}rediction model.
The innovation of CATP lies in the design of the \textbf{MW} framework, consisting of a \underline{M}anager model, multiple \underline{W}orker models, and a tailored training mechanism.
Inspired by the two-part design of GAN \cite{goodfellow2014generative} but unlike the two opposing parts in GAN, the manager and the workers are trained with competition symbiosis.
We expect each worker to perform downstream tasks in specific contexts as accurately as possible (with competition), and the manager to select the most suitable worker in the given context to achieve high overall performance (with symbiosis).
Taking CATP as an example, multiple workers need to compete with each other for training data to develop their advantages in predicting trajectories with specific moving patterns.
Meanwhile, the manager selects the best worker in the given context to perform trajectory prediction.

The MW framework can be further generalized to other tasks with context-awareness requirements.
For example, we can adopt time-series forecasting models, such as TimesNet \cite{wu2022timesnet}, as the workers to support context-aware traffic prediction.
Each worker can learn the traffic patterns in certain contexts, and the manager can select the best worker based on contextual information like time, traffic lights, accidents, and weather conditions, etc.

We conducted three quantitative experiments to comprehensively evaluate CATP model and the MW framework.
First, we compared CATP with SOTA models on context-aware trajectory prediction tasks.
The results proved that our work outperformed the SOTA models, especially when the contextual information became complicated.
Second, an ablation study further explored the performance of CATP under different model settings and training hyperparameters, to guide on training a robust CATP model.
Third, we used sequential models as the worker and compared MW-based models with SOTA models on context-aware time-series prediction tasks, which proved the generalizability of the MW framework.

In summary, the contributions of this work are as follows:
\begin{itemize}[leftmargin=10pt]
    \item We propose the novel MW framework and the tailored training mechanism, which can be generalized to handling different context-aware tasks.
    \item We implement the proposed framework into context-aware trajectory prediction and develop the CATP model.
    \item We conducted two comparative experiments and an ablation study to comprehensively evaluate the effectiveness of CATP and the generalizability of the MW framework.
\end{itemize}

%% file: content/2_related_work.tex
\section{Related Work}

This work is most relevant to the context-aware prediction of trajectory data, or more generally, that of sequential data.

\textbf{SC-based models} were end-to-end -- from contextual information and past trajectories to predicted trajectories -- with specific components that could encode the contextual information \cite{lu2022Vehicle,yu2021dynamic,fang2021mdtp,jin2022selective}.
Such models inherited the strengths and limitations of end-to-end learning \cite{glasmachers2017limits}.
A well-trained model can accurately know the impacts of contextual information on trajectories.
However, due to a lack of explainability and diagnosability, such models can be hard to train, especially when the data at both ends are complex.

Compared with these models, CATP \textbf{simplifies} the training process with the two-part model design.
Each worker only learns specific moving patterns, without interactions with the complicated contextual information.
The manager only selects the most suitable worker in the given context, rather than knowing the final trajectories.
While it may not surpass a well-tuned single context-aware model in every instance, CATP has a greater potential for achieving optimal training outcomes.

\textbf{EL-based models} trained multiple basic trajectory prediction models and applied weights on the prediction results of each model, resulting in a final prediction \cite{dong2020survey,li2022ensemble,jiang2017spatial,zhang2019prediction,wang2021johan}.
Such models benefited from advanced techniques in adapting each basic model to a subset of training data \cite{xu2022metaptp}.
However, there still exist two key issues, namely, how to split the dataset to ensure each basic model is different and how to adjust the weights.
Compared with existing methods, our method can automatically address these two issues in a data-driven manner.

For dataset partition, Jiang et al. \cite{jiang2017spatial} and Lin et al. \cite{lin2019deepstn} split the dataset from the spatial and temporal dimensions, respectively.
Instead of splitting the dataset by specific rules constructed manually, CATP utilizes the manager model to filter the most suitable data samples for training each worker model.

For model weight adjustment, Li et al. \cite{li2022ensemble} averaged the predictions of most models predicting the same direction by plurality voting.
Fazla et al. \cite{fazla2022context} transferred side information into constraints on model weights.
Le et al. \cite{le2007adaptive} continuously deleted basic models (weight is 0) with high prediction loss and added new ones (weight is 1) to adapt to the context changes.
Instead of combining the results of every base learner by weights, CATP allocates the most suitable worker for a specific trajectory prediction task.

\textbf{RL-based models} designed rewards based on contextual information and predicted the values that could gain the highest rewards \cite{co2018self,milani2023navigates}.
Thus, the results are quantifiable and thereby interpretable.

The premise of applying such models is that the subject to predict will move on the optimal path in most cases.
For example, most car drivers follow the best routes provided by navigation APPs, which will be highly rewarded by RL-based models obviously.
However, the subject may not always know which route is optimal in practice.
For example, birds may be disturbed by some environmental factors and become disoriented during migration.
Video game players may take diverse routes due to the uncertainty caused by not being able to observe the opponent's state at all times.
In these cases, it is hard for RL-based methods to determine the reward functions.
Compared to RL-based models, CATP is data-driven -- workers learn the real moving patterns in the dataset, and the manager selects workers based on their performance -- without relying on domain knowledge.

%% file: content/3_preliminaries.tex
\section{Preliminaries}

\subsection{Data Description}

\label{sec:datadesc}

Our study mainly considers two types of data, namely, trajectory data and context data.
See the glossary table in Appendix \ref{app:glossary}.

\parasum{Trajectory Data.}
A trajectory of a unit $u$ records the locations of $u$ at certain moments.
Given the current time $t$, we denote the past trajectory as $T^u_{t, \Delta t, L} = (loc^u_{t - (L-1)\Delta t}, loc^u_{t - (L - 2)\Delta t}, ..., loc^u_{t})$, where $\Delta t$ stands for the time interval between two consecutive records, and $L$ is the length of the trajectory.
Each location record $loc^u_i$ in trajectory $T^u_{t, \Delta t, L}$ is a coordinate (e.g., a 2D one can be $loc^u_i = (x^u_i, y^u_i)$).
Similarly, we denote the predicted trajectory after $t$ as $\hat{T}^u_{t + L\Delta t, \Delta t, L} = (\hat{loc}^u_{t+\Delta t}, \hat{loc}^u_{t + 2\Delta t}, ..., \hat{loc}^u_{t + L\Delta t})$, where $\hat{loc}^u_i$ is the predicted location of $u$ at time $i$.

\parasum{Context Data.}
The context data records the latest observed data frames, denoted $C_{t, \Delta t, L} = \{df_{t-(L-1)\Delta t}, df_{t-(L-2)\Delta t}, ..., df_t\}$, where each data frame $df_t$ records $N$ contextual data states (e.g., air pressure) at time $t$, denoted $df_t = \{ds^1_t, ds^2_t, ..., ds^N_t\}$.
To align different data states, we normalize each data state $ds^i_t$ in our datasets based on their value types.

\begin{itemize}[leftmargin=10pt]
    \item
    \textbf{Ranged data state} is numerical data, of which the value must be in a specific range $[MIN^i, MAX^i]$ (e.g., the time of day).
    Thus, we adopt a linear projection normalization function $(ds^i_t - MIN^i) / (MAX^i - MIN^i)$.
    \item
    \textbf{Unlimited data state} is also numerical data, of which the value starts from 0 and has no specific upper limit but only a ``soft'' upper limit $\hat{MAX}^i$ (e.g., the wind speed).
    The data value will fall within $[0,\hat{MAX}^i]$ in most cases (80\% in this work).
    We normalize the unlimited data state by $tanh(ds^i_t / \hat{MAX}^i)$, so that our model can adapt to most changes in the data state.
    \item
    \textbf{Boolean data state} is categorical data with two opposing states (e.g., daytime or nighttime).
    We simply use 0 for one state and 1 for the other.
    \item
    \textbf{Enumerated data state} is nominal data with discrete values (e.g., the bird type).
    In this work, enumerated data states with less than 10 categories are encoded by one-hot vectors.
    Others are encoded with shorter embedding vectors \cite{word2vec}, which are trained along with CATP (see Section CATP), for the sake of saving memory.
\end{itemize}

\subsection{Problem Formulation}

\input{figures/f1_sample.tex}

The problem can be formulated as, given a set $U$ of $|U|$ units and a target object $o \in U$, our model attempts to predict the future trajectory of ${o}$ based on the past trajectories of all units in $U$, with the assistance of the context data observed.
Shown as Figure \ref{fig:sample}, the input consists of two parts, namely, the observed context data $C = C_{t, \Delta t, LC}$ and a set of the past trajectories $TX = \{T^u_{t, \Delta t, LX} | u \in U\}$, where $LC$ and $LX$ are the number of data frames in context data and the length of past trajectories, respectively.
Note that we also regard the past positions of all units as important contextual information, so we record them in both $C$ and $TX$.
Given $C$ and $TX$, our model is expected to output the predicted trajectory of ${o}$ after moment $t$, denoted $\hat{TY} = \hat{T}^{o}_{t+LY\Delta t, \Delta t, LY}$, where $LY$ is the length of predicted trajectory.
The predicted trajectory $\hat{TY}$ should be as close as possible to the actual trajectory $TY = T^{o}_{t+LY\Delta t, \Delta t, LY}$.

%% file: figures/f1_sample.tex
\begin{figure}[t]
    \includegraphics[width=\linewidth]{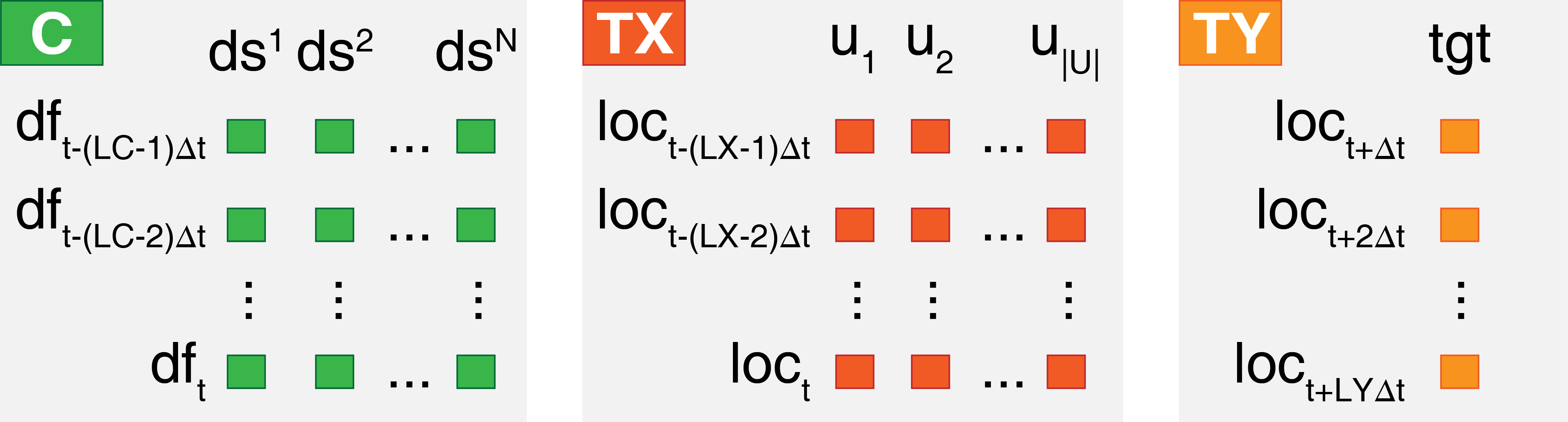}
    \caption{A data sample consists of three parts, all in matrix form. The shapes of $C$, $TX$, and $TY$ are $LC \times (N + \sum(LE^i - 1))$, $LX \times 2|U|$, and $LY \times 2$, respectively.}
    \label{fig:sample}
\end{figure}

%% file: content/4_model.tex
\section{CATP}

\label{sec:model}

This section outlines the model structure and training process of CATP, which is adaptable to models based on the MW framework for other context-aware tasks.

\subsection{Modeling}
\label{sec:arch}

\input{figures/f2_arch}

CATP consists of a manager model $M$ and $K$ worker models from $W_1$ to $W_K$, shown as Figure \ref{fig:arch}.
First, $M$ predicts the probability that each worker can accurately predict the trajectory in the context $C$.
Then, the worker with the highest probability will be selected to predict $\hat{TY}$ based on $TX$.

\subsubsection{Manager model}
The manager model performs a $K$ classification task, mapping the context data $C$ to one of the $K$ workers.
Shown as Figure \ref{fig:arch}, the manager model is built based on the well-known Transformer model \cite{vaswani2017attention,han2021transformer}, with sinusoidal positional embedding (L1) and 12 Transformer encoders (L2) by default.
There exist two customized parts as follows.
First, the embedding layer (L1) also encodes the enumerated data state in each $df_t$.
Second, following the works using Transformer for classification tasks \cite{chen2021crossvit,zhang2021token}, we add a Global Average Pooling layer (L3) and a Softmax layer (L4) to output a probability distribution over $K$ classes.

\subsubsection{Worker model}
\label{sec:worker}
The selected worker model predicts $\hat{TY}$ based on $TX$, shown as Figure \ref{fig:arch}.
Our MW framework can accommodate any sequential prediction model as the worker model.
CATP selects the Transformer model due to its excellent performance and widespread adoption across various domains \cite{han2021transformer,ren2021mtrajrec}.
The model uses sinusoidal positional embedding (L5 and L7), 12 Transformer encoders (L6) learning the moving patterns in $TX$, and 12 decoders (L8) predicting the displacement at each $\Delta t$ duration by default.
The displacement between $t + (i - 1)\Delta t$ and $t+i\Delta t$, denoted $\hat{dis}_i$, is predicted based on the start location $loc^o_t$ and previous predicted trajectory from $\hat{loc}^o_{t+\Delta t}$ to $\hat{loc}^o_{t+(i-1)\Delta t}$.
To ensure the predicted trajectory is reasonable, we add a constraint layer (L9) to limit the Euclidean distance $||\hat{dis}_i||_2$ not to exceed the maximum moving speed $MaxMS$:
\begin{equation}
    \label{eq:constraint}
    \begin{aligned}
    scale =~& Sigmoid(||\hat{dis}_i||_2) \times MaxMS~/~||\hat{dis}_i||_2, \\
    \hat{loc}^{o}_{t+i\Delta t} =~& \hat{loc}^{o}_{t+(i-1)\Delta t} + scale \times \hat{dis}_i.
    \end{aligned}
\end{equation}

\subsection{Training Mechanism}
\label{sec:train}

We propose an iterative training algorithm to train the manager and workers with competition symbiosis, shown as Figure \ref{fig:training} and Algorithm \ref{alg:train}.
\textbf{The competition} is reflected in that the workers need to compete with each other to be selected by the manager and gain more training data for advancement.
\textbf{The symbiosis} is reflected in that we use the results of the manager and workers to enhance each other to improve the overall performance of CATP.
The main training process and some details are as follows.

\subsubsection{Main training process}
At each iteration, we first train the workers, shown as the black and orange arrows in Figure \ref{fig:training} and lines 4-15 in Algorithm \ref{alg:train}.
For each data sample, the manager firstly predicts a probability $\hat{P}$ in context $C$ (arrow \circled{1}).
Then, only the worker $W_i$ with the highest probability is trained (arrow \circled{2}), to predict $\hat{TY}$ based on $TX$ (arrow \circled{3}), with $TY$ as target.
We expect to train each worker with data samples in specific contexts so that different workers can master the moving patterns in different contexts.

\input{figures/f3_training}

\input{figures/a1_training.tex}

Then, we train the manager (the black and blue arrows in Figure \ref{fig:training} and lines 16-22 in Algorithm \ref{alg:train}) to predict the probability distribution $\hat{P}$ based on $C$ (arrow \circled{1}).
The key issue is to find the target probability distribution $P$ because there exists no ground truth.
A basic idea is that the worker with a low prediction loss should have a high probability of being selected.
Thus, we use each worker to predict the trajectories (arrow \circled{3}) and obtain the prediction loss $L^W_I$ of each model $W_i$ (arrows \circled{4}-\circled{5}).
Based on each worker's loss, we can construct the simplest target probability distribution $P$ as a candidate:
\begin{equation}
    \label{eq:probOfLoss}
    \begin{aligned}
    MaxL &= \max_{1 \leq i \leq K}L^W_i,\\
    P &= Softmax(\frac{MaxL}{L^W}).
    \end{aligned}
\end{equation}

\parasum{Detail \#1: training order.}
We train the manager after the workers because the manager needs to learn about the workers' performance.
At the beginning of training, when the workers can hardly predict the trajectory, training the manager is meaningless.

\parasum{Detail \#2: training regularization.}
A key of training CATP is to balance the training of each worker.
In a bad competition, as long as a worker $W_i$ performs better than other workers, whether accurate or not, the manager tends to select it and assign more data samples to train $W_i$ at the next iteration.
In contrast, other workers, unable to improve due to lack of training data, are more difficult to be selected by the manager in the future.
Finally, only one dominant worker is well trained and selected, resulting in a lack of model diversity and context data becoming useless.

Monte-Carlo Tree Search \cite{browne2012survey} (MCTS) faces a similar issue -- whether to select the node with the highest winning rate or the ones without sufficient exploration.
Inspired by MCTS, we add a training regularization term (arrow \circled{6}, line 21, Equation \ref{eq:reg}) to consider the training volume $V_i$ -- the number of data samples used to train each worker model $W_i$ (line 15):
\begin{equation}
    \label{eq:reg}
    \begin{aligned}
        MaxV &= \max_{1 \leq i \leq K}V_i,\\
        P &= Softmax(e^{-\frac{L^W}{MaxL}} + \beta \times \frac{MaxV - V}{MaxV}),
    \end{aligned}
\end{equation}
where $e^{-L^W/MaxL}$ is the new loss term (TNewLoss), different from the one $MaxL/L^W$ in Equation \ref{eq:probOfLoss} (TOriginLoss), and $(MaxV - V)/MaxV$ is the regularization term (TReg).

For the regularization term, we expect it to be large at the beginning -- so that each worker is sufficiently trained -- and small at the end -- so that the manager can accurately learn the workers' performance.
TReg satisfies our expectation because $MaxV$ usually increases faster than $MaxV-V$, making TReg become small.
Parameter $\beta$ can also control the weight of TReg.

For the loss term, we change TOriginLoss to TNewLoss, since TReg ranges from 0 to 1, but TOriginLoss ranges from 1 to infinity, which may overshadow the impact of TReg on $\hat{P}$.
In contrast, TNewLoss has a similar range ($1/e$ to 1) to TReg, ensuring both terms can affect $\hat{P}$.

\parasum{Detail \#3: batch training.}
Figure \ref{fig:training} demonstrates how we train CATP with only one data sample, which may be easy to understand but not efficient.
Actually, our algorithm supports batch training, shown as lines 5-15 in Algorithm \ref{alg:train}.
The key is that, given a batch of data samples, we split the batch into $K$ sub-batches.
For the $i$-th sub-batch, we require that worker $W_i$ has the lowest loss on each data sample.
We further train each worker $W_i$ on the $i$-th sub-batch so that it better predicts trajectories occurring in the sub-batch.
To ensure the workers are adequately trained, we repeatedly train the workers $\alpha$ times (line 5) at each iteration.

\parasum{Detail \#4: loss function selection.}
For workers, there exist two widely used measures for evaluating the predicted trajectory, namely, the Average Displacement Error (ADE, Equation \ref{eq:ade}) and the Final Displacement Error (FDE, Equation \ref{eq:fde}).
ADE computes the average L2 distance of all pairs of predicted and ground truth locations, while FDE computes the L2 distance of the final pair of locations.
In this paper, we adopt ADE as the default loss function and denote the ADE loss of $W_i$ as $L^W_i$.
\begin{align}
    ADE(TY, \hat{TY}) &= \frac{\sum_{i=1}^{LY}\parallel \hat{loc}_{t + i\Delta t} - loc_{t + i\Delta t}\parallel_2}{LY}\label{eq:ade}\\
    FDE(TY, \hat{TY}) &= \parallel \hat{loc}_{t + LY\Delta t} - loc_{t + LY\Delta t}\parallel_2
    \label{eq:fde}
\end{align}

For the manager, we adopt Wasserstein distance \cite{arjovsky2017wasserstein} (Equation \ref{eq:wasDef}) to compute the loss of $\hat{P}$ from $P$.
Compared to KL divergence and JS divergence, Wasserstein distance is continuous and differentiable, which can provide a linear gradient and prevent collapse mode.
\begin{equation}
    \label{eq:wasDef}
    \mathbb{W}(\hat{P}, P) = \sup_{\parallel f\parallel_L \leq 1}\mathbb{E}_{x\sim \hat{P}}[f(x)] - \mathbb{E}_{x\sim P}[f(x)],
\end{equation}
where $f(x)$ is required to be a 1-Lipschitz continuous function that satisfies $|f(x_1) - f(x_2) \leq |x_1 - x_2|$, smoothing the loss function.

\subsection{Full Objective}

Our full objective is to minimize the prediction loss of the selected worker.
However, the most selected trajectory may not be necessarily selected in a specific case.
Thus, we propose top-$k$ loss -- the minimum prediction loss of $k$ workers with the highest probability.
When using top-$k$ loss, our full objective is:
\begin{equation}
    \label{eq:topkloss}
    \text{minimize~} \min_{1 \leq i \leq k}ADE(TY, W_{top_i}(TX)),
\end{equation}
where $top_i$ is the index number of the worker model with the $i$-th highest probability.

%% file: figures/f2_arch.tex
\begin{figure}[t]
    \includegraphics[width=\linewidth]{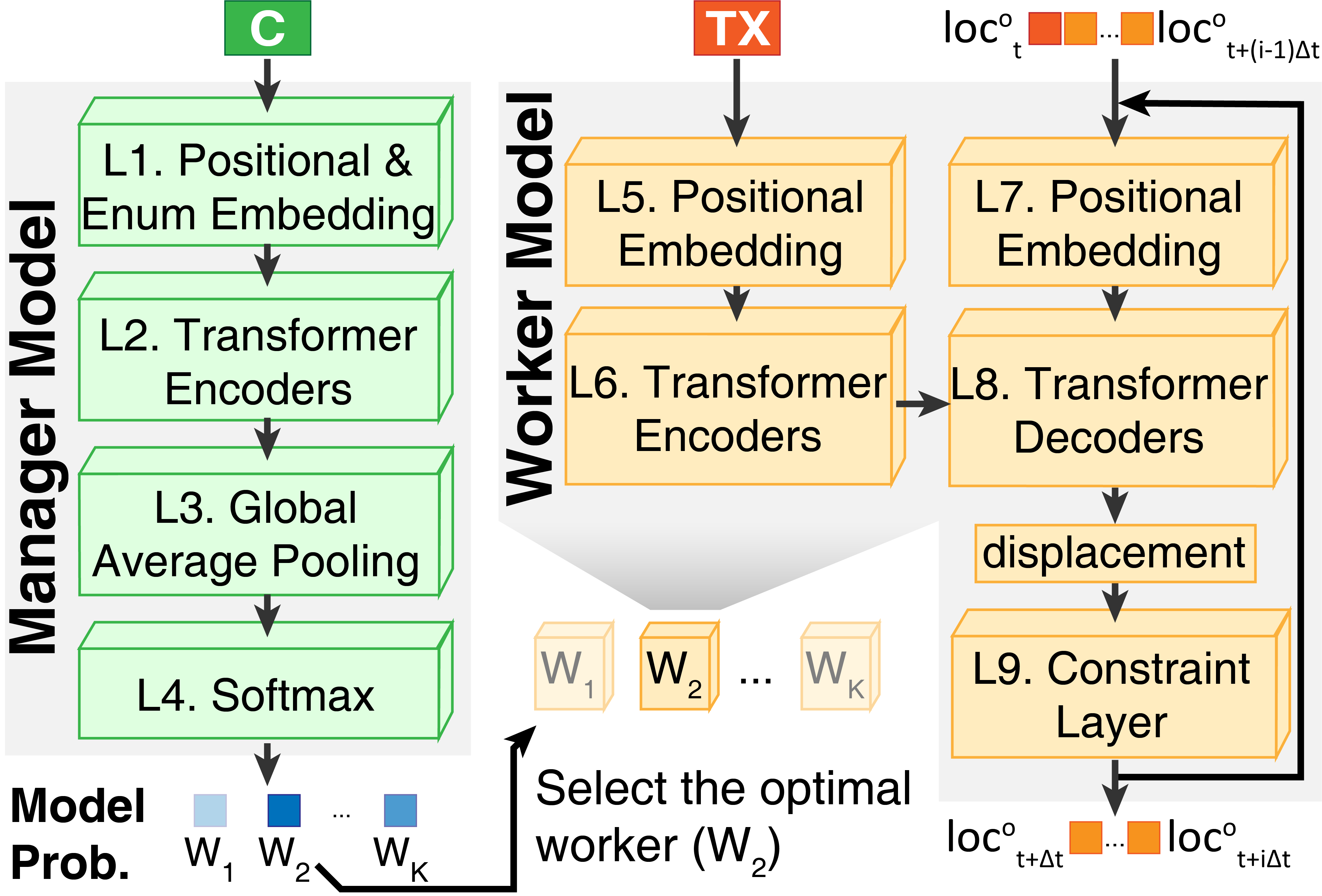}
    \caption{CATP consists of a manager model (layers L1-L4) and multiple worker models (layers L5-L9). The manager selects the optimal worker in a given context $C$, and the selected worker predicts $\hat{TY}$ based on $TX$ accurately.}
    \label{fig:arch}
\end{figure}

%% file: figures/f3_training.tex
\begin{figure}[t]
    \includegraphics[width=\linewidth]{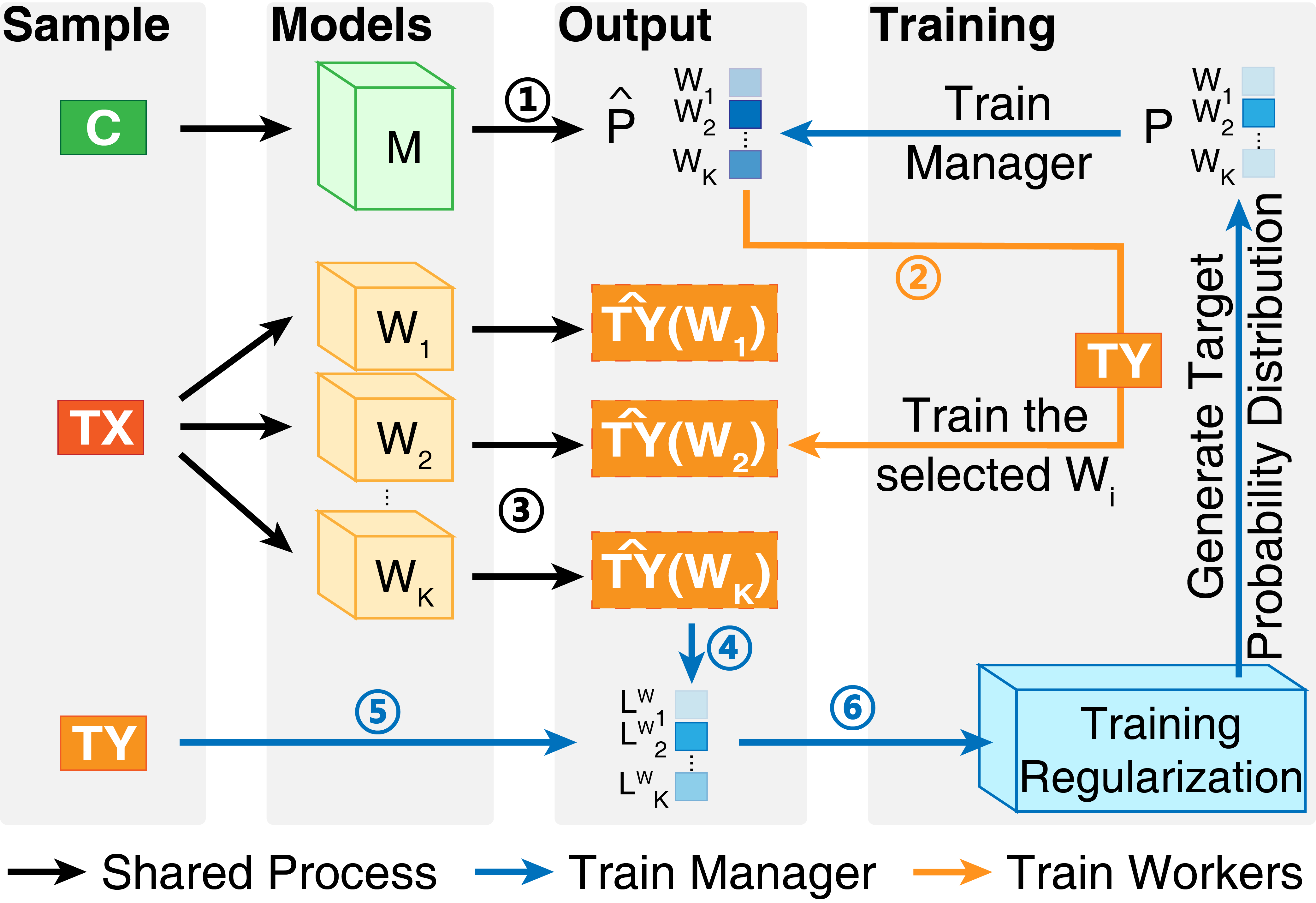}
    \caption{The proposed training mechanism. We first train workers based on the manager's results and then train the manager based on workers' performance.}
    \label{fig:training}
\end{figure}

%% file: figures/a1_training.tex
\begin{algorithm}[tb]
    \caption{The iterative training process.}
    \label{alg:train}
    \textbf{Input}: training dataset $D$, the number of workers $K$, the number of steps for training manager $T$, and batch size $b$\\
    \textbf{Output}: a manager and $K$ workers
    \begin{algorithmic}[1] %[1] enables line numbers
        % Initialize models
        \STATE $M = ManagerModel(K)$
        \STATE $W_1,...,W_K = WorkerModel() \times K$

        \FOR{number of training iterations}
            \algcmt{Fix the manager and train the workers}
            \FOR{$\alpha$ steps}
                \STATE $S = sample\_batch(D, b)$
                \STATE $c, tx, ty = parse(S)$
                \STATE $\hat{P} = M(c)$
                \FOR{each worker $W_i$}
                    \STATE $sub\_batch = []$
                    \FOR{each sample $s_j$ in $S$}
                        \IF{$\hat{P}[j][i] == max(\hat{P}[j])$}
                            \STATE $sub\_batch.append(s_j)$
                        \ENDIF
                    \ENDFOR
                    % Train W_i on S_i
                    \STATE train $W_i$ with trajectories in $sub\_batch$
                    \STATE $V_i += |sub\_batch|$
                \ENDFOR
            \ENDFOR

            \algcmt{Fix the workers and train the manager}
            \STATE $S = sample\_batch(D, b)$
            \STATE $c, tx, ty = parse(S)$
            % Find the optimal worker
            \FOR{each worker $W_i$}
                \STATE $L^W_i = ADE(W_i(tx), ty)$
            \ENDFOR
            \STATE $model\_loss = Regularize(L^W, V)$
            % Train the manager model
            \STATE train $M$ with $model\_loss$
        \ENDFOR

        \STATE \textbf{return} $M, W_1, ..., W_K$
    \end{algorithmic}
\end{algorithm}

%% file: content/5_exp.tex
\section{Evaluation}

We implemented CATP\footnote{Code for anonymous review: \projurl} and tested its performance by three quantitative experiments.
Every model in our experiments was trained in three days on a server with an NVIDIA GeForce RTX 3090 Graphics Card (24 GB) using CUDA and PyTorch \cite{pytorch}.
For each experiment, we performed a 10-round evaluation for each model over each dataset to compare model performance fairly (recorded in Appendix \ref{app:exp}).
At each round, we re-sampled the dataset and re-trained the model.

\input{content/5.1_comp1}
\input{content/5.2_ablation}
\input{content/5.3_comp2}
% \input{content/5.4_case}

%% file: content/5.1_comp1.tex
\subsection{Comparative Experiment I}
\label{sec:comp1}

In this experiment, we compared CATP with five context-aware trajectory prediction models over three datasets, to prove that CATP outperformed the SOTA context-aware trajectory prediction models.
The experimental setup and results are shown as follows.

\subsubsection{Datasets}

We prepared three datasets, listed in Table \ref{tab:dataset_traj}.
For each dataset, we put 80\%, 10\%, and 10\% of data samples in the training, validation, and test set, respectively.
\begin{itemize}[leftmargin=10pt,nosep]
    \item \texttt{DOTA} is a dataset built on 1094 public DOTA 2 games on \cite{opendota}, which records players' trajectories and 273 contextual data status (e.g., the opponents' positions).
    We chose this dataset because the abundant contextual information helped us understand the capabilities and limitations of CATP.
    Since the virtual game has complex mechanisms, we designed tailored data sampling rules to obtain data samples with good quality.
    For example, we set $LC=5$ (smaller than $LX$) because it may be difficult to learn the effects of excessive contextual information on trajectories.
    Details about the game and the dataset can be found in Appendix \ref{app:dataset}).
    \item \texttt{Bird-8} is based on an open bird migration dataset \cite{Kranstauber2022}, which records the birds' migratory trajectories in a geographic coordinate system (leading the model loss to be calculated by spherical distance) and many contextual data states, from 2008 to 2019.
    We randomly sampled 7,000 trajectories and preserved eight contextual data states most relevant to trajectory prediction in \texttt{Bird-8}, e.g., the air temperature and the wind speed (see Appendix \ref{app:dataset}).
    \item \texttt{Bird-2} and \texttt{Bird-8} are nearly identical, with only two contextual data states (i.e., the timestamp and the season).
    We set up this dataset to test our hypothesis that fewer contextual data states led to higher prediction loss.
\end{itemize}

\input{figures/t2_datasets_traj.tex}

\subsubsection{Models}
We compared four \texttt{CATP} variants with five baselines, including three SOTA ones.

\parasum{Baselines.} The five baselines are listed as follows.

\begin{itemize}[nosep,leftmargin=10pt]
    \item \texttt{LSTM} (vanilla LSTM).
    We concatenated $TX$ and $C$ at each time step and input the combined sequences into a model consisting of two LSTM layers with 128 units.
    \item \texttt{Trans} (vanilla Transformer).
    We adopted a model consisting of six transformer encoder layers and six transformer decoder layers, of which the embedding size was 128 and the input was the same as that for \texttt{LSTM}.
    \item \texttt{DATF} \cite{park2020datf} is a SOTA single context-aware trajectory prediction model, which is capable of encoding multimodal contextual information and accepts both $C$ and $TX$ as input.
    \item \texttt{Tjf} (Trajformer \cite{bhat2020trajformer}) is also a SOTA trajectory prediction model, which focuses on handling multiple agents' trajectories.
    Thus, \texttt{Tjf} accepts only $TX$ as input, which can be regarded as an advanced alternative to our worker model.
    \item \texttt{IETP} \cite{li2022ensemble} is a SOTA EL-based method for context-aware trajectory prediction.
    We set its base learners as CATP's worker model, to compare the mechanism of CATP (selecting the best worker) with that of EL (combining the results of all base learners).
    We reported the performance of \texttt{IETP} with 10 base learners (\texttt{IETP-10}) and 20 ones (\texttt{IETP-20}).
\end{itemize}

\parasum{Our model.}
We reported the performance of four \texttt{CATP} variants, varying in the number of workers (\texttt{CATP-10} or \texttt{CATP-20}) and whether to use \texttt{Tjf} as the worker (with or without \texttt{Tjf} suffixes).
Detailed settings about \texttt{CATP} variants can be found in Section \ref{sec:ablation}.

In order to ensure that all models were compared fairly, we added the same preprocessing layer (the enumerated data states embedding in our manager model) and post-process layer (the constraint layer in our worker model) to every model and trained them with a batch size of 64.
The number of epochs was 30 with proper early stopping.

\subsubsection{Results}
We reported the top-1 ADE loss (MEAN) and the standard deviation (SD) in Table \ref{tab:comparison}, which are analyzed as follows.

\textbf{CATP outperformed baselines especially when the context was complex.}
\texttt{CATP-20-Tjf} and \texttt{CATP-10} are clearly the winners on \texttt{DOTA} and \texttt{Bird-8}, respectively, where contextual information is richer.
In contrast, \texttt{DATF} performed the best on \texttt{Bird-2}, but \texttt{CATP-10} and \texttt{CATP-10-Tjf} was just slightly behind.

\input{figures/t3_comparison.tex}

\textbf{Contextual information is crucial to trajectory prediction.}
On all datasets, \texttt{LSTM}, \texttt{Trans} (which might underuse the contextual information), and \texttt{Tjf} (without $C$ as input) had high ADE losses.
Meanwhile, when predicting the bird migration trajectories, most models performed better on \texttt{Bird-8} than on \texttt{Bird-2}, which meant that the contextual information facilitated trajectory prediction.

\textbf{Models with multiple base learners (\texttt{CATP} and \texttt{IETP}) tended to outperform single context-aware models (\texttt{DATF}) when context was complex.}
\texttt{DATF} performed the best on \texttt{Bird-2}, while \texttt{CATP-10} and \texttt{IETP-20} outperformed \texttt{DATF} on others.
These results suggested that a single model might struggle to learn an effective representation of complex contextual information and understand how it affected trajectories.

\textbf{Models with competition symbiosis (\texttt{CATP}) usually outperformed traditional EL-based models (\texttt{IETP}).}
CATP models had a lower ADE loss than \texttt{IETP} on \texttt{DOTA} and \texttt{Bird-8}, respectively.
On \texttt{Bird-2}, \texttt{CATP-10} achieved a similar accuracy as \texttt{IETP}.
This indicated that the competition symbiosis enhanced CATP's performance in diverse contexts.

\textbf{CATP can empower models to be context-aware.}
The two transformer-based \texttt{CATP} models and the two with \texttt{Tjf} suffixes outperformed \texttt{Trans} and \texttt{Tjf}, respectively.
It proved that CATP could significantly improve the performance of models without context-aware designs, even SOTA ones, by automatically splitting the dataset for each worker based on contextual information.

More comparison among multiple \texttt{CATP} variants can be found in Section \ref{sec:ablation}, where we answered why \texttt{CATP-10} performed better than \texttt{CATP-20} on Bird dataset and how \texttt{Tjf} affected \texttt{CATP}'s performance as a worker.

%% file: figures/t2_datasets_traj.tex
\begin{table}[t]
    \caption{Trajectory datasets used in Section \ref{sec:comp1}. $|S|$ is the number of data samples.}
    \label{tab:dataset_traj}
    \begin{tabular}{|c|r|r|r|r|r|r|r|}
        \hline
        Datasets & $|S|$ & $|U|$ & $N$ & $\Delta t$ & $LC/LX/LY$ & $MaxMS$ \\
        \hline
        \texttt{DOTA} & $10^5$ & 10 & 273 & 0.5s & 5 / 30 / 10 & 275 \\
        \hline
        \texttt{Bird-8} & 7,000 & 1 & 8 & 300s & 30 / 30 / 10 & 4km \\
        \hline
        \texttt{Bird-2} & 7,000 & 1 & 2 & 300s & 30 / 30 / 10 & 4km \\
        \hline
        \end{tabular}
\end{table}

%% file: figures/t3_comparison.tex
\begin{table}[t]
    \caption{Results of Comparative Experiments I. We compared CATP with four baselines using the ADE loss ($mean\pm sd$).}
    \label{tab:comparison}
    \begin{tabular}{c|r@{}r@{}r|r@{}r@{}r|r@{}r@{}r}
        \toprule
        \multirow{2}{*}{Model} & \multicolumn{9}{l}{Dataset}\\
        \cmidrule(lr){2-10}
        & \multicolumn{3}{l|}{\texttt{DOTA}} & \multicolumn{3}{l|}{\texttt{Bird-8}/km} & \multicolumn{3}{l}{\texttt{Bird-2}/km} \\
        \midrule
        \cmpr{LSTM}{1279}{199}{21.22}{3.06}{23.41}{4.63}
        \cmpr{Trans}{1227}{128}{20.68}{2.76}{24.42}{2.10}
        \cmpr{Tjf}{969}{100}{19.43}{1.95}{19.43}{1.95}
        \cmpr{DATF}{811}{116}{15.90}{0.89}{\best{16.64}}{\best{0.86}}
        \cmpr{IETP-10}{625}{120}{16.44}{0.81}{20.27}{1.44}
        \cmpr{IETP-20}{556}{49}{13.78}{1.12}{17.77}{1.64}
        \midrule
        \cmpr{CATP-10}{563}{80}{\best{11.94}}{\best{0.45}}{17.74}{2.83}
        \cmpr{CATP-20}{426}{\best{33}}{17.08}{1.41}{23.31}{2.58}
        \cmpr{CATP-10-Tjf}{538}{59}{\best{11.94}}{1.01}{17.26}{2.40}
        \cmpr{CATP-20-Tjf}{\best{423}}{39}{18.35}{1.80}{21.22}{3.36}
        \bottomrule
        \end{tabular}
\end{table}

%% file: content/5.2_ablation.tex
\subsection{Ablation Study}
\label{sec:ablation}

\input{figures/t6_ablation.tex}

We conducted this study to better know the impacts of each key hyperparameter on CATP's performance, thereby guiding how to build and train a robust CATP model.

\subsubsection{Training Traps}
Before analyzing the detailed results of each experimental group, we prefer to introduce two common training traps of CATP that occurred in many experimental groups.
When CATP falls into these two traps, the manager model cannot work properly, leading the whole model to almost degenerate into a vanilla Transformer model.
\begin{enumerate}[leftmargin=10pt,label={\bf T{{\arabic*}}}]
    \item \label{trap:single}
    \textbf{One dominant worker was used in all contexts}, marked by $\curlywedge$ in Table \ref{tab:ablation}.
    When the regularization term had little effect on the training process, the manager tended to only select the worker with the lowest ADE loss and train it only, using it in all contexts.
    In this case, the manager's accuracy is higher than 90\%.
    \item \label{trap:equal}
    \textbf{All workers were trained equally}, marked by $\thickapprox$ in Table \ref{tab:ablation}.
    When the regularization term had an overweight effect on the training process, the manager tended to train each worker equally.
    Therefore, each worker could not learn unique moving patterns, and the manager could not know the suitable context for each worker, leading the manager's accuracy to be lower than 20\%.
\end{enumerate}

The manager's selection indicates what happened during the training process, helping model developers fine-tune the hyper-parameters.
For example, when the manager always selects the same worker, we can enhance the weight of the regularization term.

\subsubsection{Results}
After knowing the two training traps above, we elaborated the analysis of each experimental group as follows.

\textbf{Group A: a large $K$ did not always result in lower prediction losses.}
We changed $K$ and $\alpha$ correspondingly so that the number of data samples used to train each worker remains roughly the same.
Due to the training regularization term, the manager model tends to use all worker models.
However, when the number of worker models is small, one worker needs to learn many moving patterns, leading to high losses.
When the number of worker models is large, many workers learn similar moving patterns and compete for the same training data, resulting in insufficient training data.

\textbf{Group B: competition symbiosis helps improve model performance.}
We compared training algorithms with and without competition symbiosis (CSTrain) to examine its effects.
For the one without CSTrain, we split the dataset randomly into $K$ subsets, each for one worker.
After all the workers finished training, we trained the manager model based on the workers' performance.
Without CSTrain, the workers were trained equally, leading CATP to fall into Trap \ref{trap:equal}.

\textbf{Group C: $\alpha$ should be fine-tuned during training.}
When $\alpha=1$, each worker obtained insufficient training data at each iteration and could not quickly converge at the beginning of training.
Meanwhile, since every worker has a large loss, the manager model mainly relied on the regularization term, falling into Trap \ref{trap:equal}.
A large $\alpha$ makes workers converge quickly but eventually causes them to overfit.
It is better to start with a large $\alpha$ and keep reducing it.

\textbf{Group D: $\beta$ is crucial for avoiding the two traps.}
When $\beta$ was zero, the manager model completely relies on the loss, falling into Trap \ref{trap:single}.
In contrast, when $\beta$ was one, the manager model relied more on the training counts, falling into Trap \ref{trap:equal}.

\textbf{Group E: appropriate regularization functions help avoid Trap \ref{trap:single}.}
We tested two more regularization functions.
One was Equation \ref{eq:probOfLoss}, which had no training regularization term.
The other was Equation \ref{eq:noNorm}, which directly added a regularization term to Equation \ref{eq:probOfLoss}, without normalizing the loss term into $[0,1]$.
Both of these two regularization functions led CATP to fall into Trap \ref{trap:single}.

\begin{equation}
    \label{eq:noNorm}
    P = Softmax(\frac{MaxL}{L^W} + \beta \times \frac{MaxV - V}{MaxV})
\end{equation}

\textbf{Group F: loss functions affect model accuracy.}
We tested cross entropy and FDE as the loss functions for training the manager model and worker models, respectively.
Cross entropy underperformed Wasserstein loss, probably because Wasserstein loss allows the manager to converge quickly and consistently.
FDE loss undoubtedly caused a higher ADE loss, but also a lower manager accuracy, probably because two moving strategies with the same destination, but different trajectories were regarded as the same.
Thus, many workers had to learn similar moving patterns, similar to the issue when setting $K$ to 30 in Group A.

\textbf{Group G: CATP supports continuous performance improvement with SOTA model integration.}
We replaced each worker model with \texttt{Tjf} and observed lower losses and a higher manager accuracy.
It proved that workers' better performance can benefit the training of the manager and the overall performance.

%% file: figures/t6_ablation.tex
\begin{table*}[t]
    \caption{Results of ablation study.
    There existed two control groups (\texttt{CATP-10} and \texttt{CATP-20}) and seven experimental groups (from A to G), varying in the number of worker models ($K$), whether to train with competition symbiosis (CSTrain), training parameters ($\alpha$ and $\beta$), the regularization function (Reg.), the loss functions (LossM and LossW), and the worker models (Worker).
    The empty cells were filled with the same settings of \texttt{CATP-20}.
    These model variations were evaluated by the top-1/top-3 ADE loss ($MEAN\pm SD$) and manager accuracy (MngAcc, whether the top 3 workers had the lowest loss among all workers).
    }
    \label{tab:ablation}
    \begin{tabular}{c|cccccccc|r@{}r@{}rr@{}r@{}rr@{}l}
        \toprule
        Group&$K$&CSTrain&$\alpha$&$\beta$&Reg.&LossM&LossW&Worker&\multicolumn{3}{c}{Top-1 ADE}&\multicolumn{3}{c}{Top-3 ADE}&\multicolumn{2}{c}{MngAcc}\\

        \ablt{\texttt{CATP-20}}{1}
        \ablr{\Checkmark}{20}{100}{0.1}{Eq. \ref{eq:reg}}{Wasserstein}{ADE}{Transformer-based}{426}{33}{375}{28}{81}{}

        \ablt{\texttt{CATP-10}}{1}
        \ablr{\Checkmark}{10}{50}{}{}{}{}{}{564}{81}{483}{65}{83}{}

        \ablt{A}{2}
        &1&&5&&&&&&1228&$\pm$&129&&/&&100\%&$\curlywedge$\\
        % \ablr{}{1}{5}{}{}{}{}{}{1228}{129}{/}{}{100}{$\curlywedge$}
        \ablr{}{30}{150}{}{}{}{}{}{453}{28}{392}{25}{75}{}

        \ablt{B}{1}
        \ablr{\XSolidBrush}{}{}{}{}{}{}{}{1104}{106}{1029}{91}{8}{$\thickapprox$}

        \ablt{C}{2}
        \ablr{}{}{1}{}{}{}{}{}{983}{82}{910}{77}{10}{$\thickapprox$}
        \ablr{}{}{500}{}{}{}{}{}{554}{47}{521}{40}{86}{}

        \ablt{D}{2}
        \ablr{}{}{}{0}{}{}{}{}{1054}{91}{958}{83}{93}{$\curlywedge$}
        \ablr{}{}{}{1}{}{}{}{}{879}{72}{844}{67}{12}{$\thickapprox$}

        \ablt{E}{2}
        \ablr{}{}{}{}{Eq. \ref{eq:probOfLoss}}{}{}{}{941}{124}{883}{105}{98}{$\curlywedge$}
        \ablr{}{}{}{}{Eq. \ref{eq:noNorm}}{}{}{}{793}{110}{714}{96}{94}{$\curlywedge$}

        \ablt{F}{2}
        \ablr{}{}{}{}{}{Cross Entropy}{}{}{603}{53}{539}{48}{62}{}
        \ablr{}{}{}{}{}{}{FDE}{}{502}{47}{474}{39}{77}{}

        \ablt{G}{1}
        \ablr{}{}{}{}{}{}{}{Trajformer \cite{bhat2020trajformer}}{423}{39}{362}{23}{83}{}

        \bottomrule
    \end{tabular}
\end{table*}

%% file: content/5.3_comp2.tex
\subsection{Comparative Experiment II}
\label{sec:comp2}

This experiment was conducted to prove the generalizability of the MW framework.
We adopted time-series prediction models as the workers to conduct context-aware time-series prediction tasks.
The experimental setup and results are as follows.

\subsubsection{Datasets}
We tested our model on two open datasets, listed as follows and in Table \ref{tab:dataset_ts}.
\begin{itemize}[leftmargin=10pt,nosep]
    \item \texttt{SP}\footnote{Solar Power: https://www.nrel.gov/grid/solar-power-data.html} is a dataset recording solar power data in 2006, with six contextual variables.
    We trained the models using the data in California and New York to ensure different changing patterns.
    \item \texttt{ET}\footnote{ET: https://github.com/zhouhaoyi/ETDataset} is a dataset used as a benchmark by many recent models.
    We trained the models on the subset named ``ETT-h1'' because of its rich contextual information.
    We set oil temperature (OT) as the target channel and used the other eight channels as the contexts.
\end{itemize}

To obtain sufficient data samples, we used different lengths of stride and time windows for data sampling, depending on the total length of the time series.
The data sampling proportions in the training, validation, and test sets are 80\%, 10\%, and 10\% respectively.

\input{figures/t4_datasets_ts.tex}

\subsubsection{Models}
We compared three \texttt{CATP} variants with six time-series prediction models, including four SOTA ones.
All these models were trained with a batch size of 32.
The number of epochs was set to 8 with proper early stopping.

\parasum{Baselines.}
The first two baselines were \texttt{LSTM} and \texttt{Trans}, which were also used in Comparative Experiment I.
We also selected four SOTA models -- \texttt{Informer} \cite{zhou2021informer}, \texttt{Autoformer} \cite{wu2021autoformer}, \texttt{FEDFormer} \cite{zhou2022fedformer}, and \texttt{TimesNet} \cite{wu2022timesnet}.
Since these baselines have no components for contextual channels, we concatenated the contextual channels and the target channel to construct multivariate time-series data as the input of these baseline models.
Based on such considerations, we did not select models designed to handle univariate time-series data as our baselines (e.g., ARIMA \cite{ariyo2014stock} and DeepAR \cite{salinas2020deepar}).

\parasum{Our model}. We replaced the workers with time-series prediction models, namely, \texttt{LSTM}, \texttt{Trans}, and \texttt{TimesNet}, and built \texttt{MW-LSTM}, \texttt{MW-Trans}, and \texttt{MW-TN}, respectively.
In contrast to the baselines, we used the contextual channels and the to-be-predicted channel as the input of manager and workers, respectively, so that we can highlight the capabilities and limitations of the MW framework.
Considering the complexity of each dataset, we set the number of workers to 10, 20, and 15 for \texttt{MSL}, \texttt{SP}, and \texttt{ET}, respectively.
Other hyperparameters were set to the optimal values tested in our Ablation Study (Section \ref{sec:ablation}).

\subsubsection{Results}

We reported the top-1 MSE loss ($MSE = \frac{1}{n}\sum^n_{i=1}(y - \bar{y})^2$) in Table \ref{tab:comparison2}, which proved that the MW framework could be applied to other context-aware tasks.
Here are detailed analysis.
\textbf{1) The MW framework could significantly improve the performance of simple models.} \texttt{MW-LSTM} and \texttt{MW-Trans} ourperformed \texttt{LSTM} and \texttt{Trans} and achieved a similar accuracy as \texttt{Informer}.
\textbf{2) The MW framework and SOTA models were mutually beneficial.} \texttt{MW-TN} achieved better performance than both \texttt{TimesNet} and other \texttt{MW}-based models. The framework provided a better training set for SOTA models, while SOTA models enhanced the performance of the model as a whole.
% 3) \textbf{A large training set facilitated CATP's performance.} Compared with \texttt{TimesNet}, \texttt{CATP-TN} achieved a \dyu{slightly} larger improvement on \texttt{SP} than \texttt{ET}.
% Since CATP's training mechanism splits the training set into $K$ subsets, a large training set helps ensure that each worker is sufficiently trained.

\input{figures/t5_comparison2.tex}

%% file: figures/t4_datasets_ts.tex
\begin{table}[t]
    \caption{Two time-series datasets used in Section \ref{sec:comp2}. We obtained $|S|$ data samples from $Num$ original temporal sequences with length $Len$, at a certain Stride.}
    \label{tab:dataset_ts}
    \begin{tabular}{|c|r|r|r|r|r|}
        \hline
        Datasets & $Len$ $\times$ $Num$ & Stride & |S| & $N$ & $LC/LX/LY$ \\
        \hline
        % \texttt{MSL} & 2852 & 3 & 935 & 24 & 24/24/24 \\
        % \hline
        % \texttt{M4} & 2,794 $\times$ 10,016 & 347 & 90,144 & 24 & 10/10/8 \\
        % \hline
        \texttt{SP} & 105,120 $\times$ 2 & 12 & 17,490 & 5 & 96/96/96 \\
        \hline
        \texttt{ET} & 17,420 $\times$ 1 & 10 & 1,731 & 8 & 60/60/60 \\
        \hline
        \end{tabular}
\end{table}

%% file: figures/t5_comparison2.tex
\begin{table}[t]
    \caption{Results of Comparative Experiment II. We compared our model with six baselines using the MSE loss (MEAN $\pm$ SD).}
    \label{tab:comparison2}
    % \begin{tabular}{c|r@{}r@{}r|r@{}r@{}r|r@{}r@{}r}
    \begin{tabular}{c|r@{}r@{}r|r@{}r@{}r}
        \toprule
        % \multirow{2}{*}{Model} & \multicolumn{9}{l}{Dataset}\\
        \multirow{2}{*}{Model} & \multicolumn{6}{l}{Dataset}\\
        % \cmidrule(lr){2-10}
        \cmidrule(lr){2-7}
        % & \multicolumn{3}{l|}{\texttt{MSL}} & \multicolumn{3}{l|}{\texttt{SP}} & \multicolumn{3}{l}{\texttt{ET}} \\
        & \multicolumn{3}{l|}{\texttt{SP}} & \multicolumn{3}{l}{\texttt{ET}} \\

        \midrule

        \cmptsr{LSTM}{0.32}{0.04}{5.69}{0.54}{1.40}{0.12}
        \cmptsr{Trans}{0.34}{0.05}{5.57}{0.63}{1.23}{0.13}
        \cmptsr{Informer}{0.27}{0.03}{4.81}{0.81}{1.04}{0.12}
        \cmptsr{Autoformer}{0.20}{0.02}{3.16}{0.31}{0.50}{0.05}
        \cmptsr{FEDFormer}{0.19}{0.02}{3.16}{0.32}{0.46}{0.05}
        \cmptsr{TimesNet}{\best{0.16}}{0.02}{2.93}{0.23}{0.47}{0.05}

        \midrule

        \cmptsr{MW-LSTM}{0.25}{0.03}{4.86}{0.39}{0.99}{0.10}
        \cmptsr{MW-Trans}{0.26}{\best{0.01}}{4.74}{0.56}{1.00}{0.11}
        \cmptsr{MW-TN}{0.23}{0.03}{\best{2.86}}{\best{0.13}}{\best{0.45}}{\best{0.04}}
        \bottomrule
        \end{tabular}
\end{table}

% \begin{table}[t]
%     \caption{Results of Comparative Experiment II. We compared CATP with six baselines using the MSE loss (MEAN $\pm$ SD).}
%     \label{tab:comparison2}
%     \begin{tabular}{c|r@{}r@{}r|r@{}r@{}r}
%         \toprule
%         \multirow{2}{*}{Model} & \multicolumn{6}{l}{Dataset}\\
%         \cmidrule(lr){2-7}
%         & \multicolumn{3}{l|}{\texttt{SP}} & \multicolumn{3}{l}{\texttt{ET}} \\

%         \midrule

%         \cmptsr{LSTM}{0.32}{0.04}{5.69}{0.54}{1.40}{0.12}
%         \cmptsr{Trans}{0.34}{0.05}{5.57}{0.63}{1.23}{0.13}
%         \cmptsr{Informer}{0.27}{0.03}{4.81}{0.81}{1.04}{0.12}
%         \cmptsr{Autoformer}{0.20}{0.02}{3.16}{0.31}{0.50}{0.05}
%         \cmptsr{FEDFormer}{0.19}{0.02}{3.16}{0.32}{0.46}{0.05}
%         \cmptsr{TimesNet}{\best{0.16}}{0.02}{2.93}{0.23}{0.47}{0.05}

%         \midrule

%         \cmptsr{CATP-LSTM}{0.25}{0.03}{4.86}{0.39}{0.99}{0.10}
%         \cmptsr{CATP-Trans}{0.26}{\best{0.01}}{4.74}{0.56}{1.00}{0.11}
%         \cmptsr{CATP-TN}{0.23}{0.03}{\best{2.86}}{\best{0.13}}{\best{0.45}}{\best{0.04}}
%         \bottomrule
%         \end{tabular}
% \end{table}

%% file: content/6_discussion.tex
\section{Discussion}

\parasum{Generalizability.}
Although CATP mainly focuses on trajectory prediction, the MW framework -- the core contributions of our research -- can be generalized to many context-aware tasks.
The MW framework essentially provides a data-driven method to automatically split the training dataset based on contextual factors and workers' performances.
Such a method uses the manager model to perceive contextual information and allows the replaceable workers to focus only on downstream tasks.
Meanwhile, using SOTA models as the worker could enhance the overall model performance, resulting in continuous improvement on CATP as the SOTA models evolve.

\parasum{Why the MW framework.}
Increasingly dynamic and complex contextual information results in inefficiencies for a single model to fully leverage such information.
As outlined in our Introduction and Related Work sections, ensemble methods are outperforming single-model approaches.
The main challenge with ensemble methods is the considerable effort and expertise required to partition training samples so that the ``workers'' can effectively learn the intended knowledge.
Our solution automates this process, enabling the ``manager''  to allocate the most appropriate training data to each worker and harness the potential of existing sequential models.
The experimental results confirm our design's efficacy.

\parasum{Limitations and future works.}
The number of worker models is highly dependent on the variety of trajectories to predict in different contexts.
We usually need to fine-tune this hyperparameter and re-train the whole model when the trajectory patterns change (e.g., when applying to a new application).
In the future, we plan to study how to automatically and progressively fine-tune the number of workers.
One possible solution is to allow the manager to retire underperforming workers and bring in new ones during training.
It is also a promising solution to use some pre-trained models as the workers and allow the manager to fine-tune them quickly.

%% file: content/7_conclusion.tex
\section{Conclusion}

We introduced a context-aware trajectory prediction model, which could effectively use complex contextual information to improve prediction accuracy.
Similar to the GAN networks, our proposed model consisted of several worker models for predicting trajectory and a manager model for selecting the best worker model in the given context.
We further proposed a training mechanism with competition symbiosis, where the results of the manager and the workers could be used to train each other, and the workers with lower prediction losses can gain more training data to perform better in specific contexts.
We conducted two comparative experiments to evaluate the performance and generalizability of CATP, where CATP effectively improved the prediction accuracy of both trajectory and time-series data, especially when the context data is complex.
An ablation study further evaluated the performance of several model variations and provided guidance on how to train a robust CATP model.
% A case study further illustrated the prediction results and interpretability of CATP.

%% file: meta/acks.tex
% \begin{acks}
%     To Robert, for the bagels and explaining CMYK and color spaces.
% \end{acks}

%% file: content/a1_glossary_table.tex
\pagebreak

\section{Glossary Table}
\label{app:glossary}

\begin{table}[H]
    \caption{Glossary Table.}
    \begin{tabular}{p{0.4in}p{4in}p{1.2in}}
    \toprule
    Symbol & Description & Note \\

    \midrule
    \multicolumn{3}{l}{\cellcolor[gray]{0.9} Input and Output} \\

    \cmidrule(lr){1-3}
    \gtr{$TX$}{$\{T^{u}_{t, \Delta t, LX} | u \in U\}$}{the past trajectories of all recorded game units in $U$}
    \gtr{$C$}{$C_{t, \Delta t, LC}$}{the context data}
    \gtr{$TY$}{$T^p_{t+LY\Delta t, \Delta t, LY}$}{the real trajectory of the player to predict after $t$}
    \gtr{$\hat{TY}$}{$\mathbb{\hat{T}}^p_{t, \Delta t, LY}$}{the predicted trajectory of the player to predict after $t$}

    \midrule
    \multicolumn{3}{l}{\cellcolor[gray]{0.9} Data-Related Symbols} \\

    \cmidrule(lr){1-3}  % time
    \gtr{$t$}{}{a time point}
    \gtr{$\Delta t$}{1 sec}{the time interval of data capturing}

    \cmidrule(lr){1-3}  % context
    \gtr{$C_{t, \Delta t, L}$}{$(df_{t-(L-1)\Delta t}, ..., df_t)$}{the context data consisting of the latest $L$ data frames observed by $p$}
    \gtr{$df_t$}{$(ds^1_t, ..., ds^N_t)$}{the data frame when observing the game at time $t$, consisting of $N$ data states}
    \gtr{$ds^i_t$}{}{the $i$-th data state in data frame $df_t$}

    \cmidrule(lr){1-3}  % trajectory
    \gtr{$T^u_{t, \Delta t, L}$}{$(loc^u_{t - (L-1)\Delta t}, ..., loc^u_{t})$}{the trajectory of $u$ consisting of the latest $L$ locations}
    \gtr{$\hat{T}^u_{t+L\Delta t, \Delta t, L}$}{$(loc^u_{t+\Delta t}, ..., loc^u_{t + L\Delta t})$}{the predicted trajectory of $u$ consisting of the next $L$ locations}
    \gtr{$u \in U$}{}{a set of game units whose trajectories are recorded}
    \gtr{$p$}{$p \in U$}{the player to predict}
    \gtr{$loc^u_t$}{$(x^u_t, y^u_t)$}{the 2D coordinate of $u$ at time $t$}

    \cmidrule(lr){1-3}  % sample
    \gtr{$s$}{$(C, TX, TY)$}{a sample}
    \gtr{$S$}{$(s_1, ..., s_{|S|})$}{a batch of samples}
    \gtr{$LC$}{5}{the number of data frames in context data}
    \gtr{$LX$}{30}{the length of past trajectories}
    \gtr{$LY$}{10}{the length of the trajectories to predict}

    \midrule
    \multicolumn{3}{l}{\cellcolor[gray]{0.9} Model-Related Symbols}    \\

    \cmidrule(lr){1-3}  % model architecture
    \gtr{$K$}{20}{the number of worker models}
    \gtr{$k$}{3}{the top-$k$ worker models used to evaluate the model}
    \gtr{$L$}{}{the loss of the whole model}

    \cmidrule(lr){1-3}  % manager model
    \gtr{$M$}{}{the manager model}
    \gtr{$\hat{P}$}{}{the probability of each model selected by the manager model}
    \gtr{$L^M$}{}{the loss of the manager model}

    \cmidrule(lr){1-3}  % worker model
    \gtr{$W_i$}{}{the $i$-th worker model}
    \gtr{$V_i$}{}{the training volume of $W_i$}
    \gtr{$L^W_i$}{}{the loss of the $i$-th worker model}
    \gtr{$\hat{TY}_i$}{}{$\hat{TY}$ of the $i$-th worker model}
    \bottomrule
    \end{tabular}
\end{table}

%% file: content/a2_data.tex
\pagebreak

\section{Datasets}

\label{app:dataset}

\subsection{DOTA}

\subsubsection{Introduction to the Game}

DOTA 2 is a typical MOBA (Multiplayer Online Battle Arena) game. Two teams, each with five players, need to compete against each other on a well-designed map. To achieve effective teamwork in a highly competitive game, the players usually need to make competition plans based on the game context, and the competition plans require players to move to certain places on the map. For example, when an allied tower (a type of building) is about to be destroyed by the opponents (the contextual information), the players may move to the tower to protect it (the trajectories). A full introduction can be found on the wiki page: https://dota2.fandom.com/.

\textbf{Why this dataset?}
Video games are becoming a new arena for researchers to gain a better understanding of the capabilities and limitations of AI, for many reasons:
\begin{itemize}[leftmargin=10pt]
    \item \parasum{High Data Volume}. A popular video game, such as PUBG and LOL, may have tens of millions of players. These players can generate a vast amount of game data every day. These data can be adequate for training a deep neural network optimally. We can get access to public data on OpenDOTA \cite{opendota}, an open-source DOTA 2 data platform sponsored by OpenAI.
    \item \parasum{High Quality}. For many video games, the data is stored in a well-organized data structure. Except for rare cases of game bugs, the data can be considered correct and accurate. We can hardly find missing or misrecorded values, which may be common in a real-world dataset.
    \item \parasum{High Completeness}. For a bird migration dataset, we may find much contextual information not recorded. For example, the birds may consider where there is a lake so that they can drink water. However, we cannot guarantee that all bird migration datasets record the position of each water source along the migration route. In contrast, the video game dataset is complete. Almost all the information that motivates the players' movements is recorded, except their mental activity. Thanks to this point, we recorded hundreds of contextual data states in our video game dataset, which greatly tested the capabilities of our model.
\end{itemize}

\subsubsection{Data Description}

We provide our code for downloading and preprocessing the dataset in our code repository.
Finally, our dataset considered 99 game units and 273 contextual data states, shown as follows.

\parasum{Game Units.}
Shown as Table \ref{tab:u}, the set $U$ includes 99 game units in total.
Note that although some game units are static, we record their trajectories as POIs to improve the interpretability of our model, inspired by Bartoli et al. \cite{bartoli2018context} who involving static obstacles into the trajectory prediction of pedestrian.

\begin{table}[h]
    \caption{All the game units in $U$.}
    \label{tab:u}
    \begin{tabular}{p{1.6in}p{0.6in}p{1in}}
        \toprule
        Game Unit & Counts & Static or Moveable \\
        \midrule
        Heroes controlled by players & 2 $\times$ 5 & \multirow{4}{*}{Moveable} \\
        Couriers & 2 $\times$ 5 & \\
        Lane Creeps (at the front) & 2 $\times$ 3 & \\
        Roshan & 1 & \\
        \cmidrule(lr){1-3}
        Towers & 2 $\times$ 11 & \multirow{7}{*}{Static} \\
        Barracks & 2 $\times$ 6 & \\
        Base & 2 & \\
        Outposts & 2 & \\
        Runes & 4 & \\
        Neutral Creeps Camps & 16 & \\
        Ward Spots & 14 & \\
        \bottomrule
    \end{tabular}
\end{table}

\pagebreak

\parasum{Contextual data states.}
Shown as Table \ref{tab:dds}, each data frame includes 283 data states.
According to how players observe these data states, we classify them into three types, which determine how we record the value.

\begin{table}[H]
    \caption{All the data states considered as context data.}
    \label{tab:dds}
    \begin{tabular}{p{0.6in}p{1.6in}p{1.3in}p{3in}}
    \toprule
    Value Types & Data States & Normalization Params & Note \\
    \midrule

    \multicolumn{4}{l}{\cellcolor[gray]{0.9}Consensus data states are always known by all players} \\
    \cmidrule(lr){1-4}
    \multirow{2}{*}{\vta}
    \dsr{38 $\times$ Health of buildings}{MIN=0, MAX=3000}{The health points of different buildings are also different. However, all are below 3000.}
    \dsr{}{}{}

    \cmidrule(lr){1-4}
    \multirow{2}{*}{\vtb}
    \dsr{Current time}{$\hat{MAX}$=2400s}{}
    \cmidrule(lr){2-4}
    \dsr{10 $\times$ Respawn time of heroes}{$\hat{MAX}$=60s}{}

    \cmidrule(lr){1-4}
    \multirow{3}{*}{\vtc}
    \dsr{Circadian state}{0: daytime, 1: nighttime}{}
    \cmidrule(lr){2-4}
    \dsr{10 $\times$ Survival states of heroes}{0: alive, 1: dead}{}
    \cmidrule(lr){2-4}
    \dsr{38 $\times$ Survival states of buildings}{0: alive, 1: dead}{}
    \cmidrule(lr){2-4}
    \dsr{2 $\times$ Which team controls the outposts}{0: Radiant, 1: Dire}{}

    \cmidrule(lr){1-4}
    \multirow{2}{*}{\vtd}
    \dsr{10 $\times$ Heroes picked}{embedding vector, $LE=5$}{}
    \dsr{10 $\times$ Roles}{one-hot vector, $LE=10$}{}
    \midrule

    \multicolumn{4}{l}{\cellcolor[gray]{0.9}Vision-based data states are updated only when they can be observed by $p$}    \\
    \cmidrule(lr){1-4}
    \multirow{3}{*}{\vta}
    \dsr{10 $\times$ Current positions of heroes}{MIN=(8240,8220), MAX=(24510,24450)}{Actually, a position can be regarded as two ranged numbers. For convenience, we regard it as one data state.}
    \cmidrule(lr){2-4}
    \dsr{10 $\times$ Current positions of couriers}{same as above}{}
    \cmidrule(lr){2-4}
    \dsr{10 $\times$ Positions of wards}{same as above}{}
    \cmidrule(lr){2-4}
    \dsr{10 $\times$ Levels of heroes}{MIN=1,MAX=30}{}

    \cmidrule(lr){1-4}
    \multirow{4}{*}{\vtb}
    \dsr{10 $\times$ Health of heroes}{$\hat{MAX}$=3000}{}
    \cmidrule(lr){2-4}
    \dsr{10 $\times$ Mana of heroes}{$\hat{MAX}$=1700}{}
    \cmidrule(lr){2-4}
    \dsr{10 $\times$ Gold of heroes}{$\hat{MAX}$=15000}{}
    \cmidrule(lr){2-4}
    \dsr{Health of Roshan}{$\hat{MAX}$=4000}{}

    \cmidrule(lr){1-4}
    \multirow{2}{*}{\vtc}
    \dsr{2 $\times$ Existence of power runes}{0: without, 1: with}{}
    \dsr{}{}{}

    \cmidrule(lr){1-4}
    \multirow{2}{*}{\vtd}
    \dsr{2 $\times$ Power rune types}{one-hot vector, $LE=7$}{}
    \cmidrule(lr){2-4}
    \dsr{60 $\times$ Items purchased}{embedding vector, $LE=7$}{}
    \midrule

    \multicolumn{4}{l}{\cellcolor[gray]{0.9}Knowledge-based data states are updated when the data states are observable and can be known by players using their knowledge} \\
    \cmidrule(lr){1-4}
    \multirow{2}{*}{\vta}
    \dsr{6 $\times$ Positions of lane creeps}{same as the positions mentioned above}{Knowledge: The lane creeps move on certain paths.}
    \cmidrule(lr){2-4}
    \dsr{Position of Roshan}{same as above}{Knowledge: Roshan will be respawned at a fixed place.}

    \cmidrule(lr){1-4}
    \multirow{4}{*}{\vtc}
    \dsr{Survival state of Roshan}{0: alive, 1: dead}{Knowledge: There will be a broadcast when it is dead, and it must be respawned 12 minutes after its last death.}
    \cmidrule(lr){2-4}
    \dsr{2 $\times$ Existence of bounty runes}{0: without, 1: with}{Knowledge: Every three minutes}
    \cmidrule(lr){2-4}
    \dsr{2 $\times$ Existence of water runes}{0: without, 1: with}{Knowledge: At the 2nd and 4th minutes}
    \cmidrule(lr){2-4}
    \dsr{16 $\times$ Existence of neutral creeps}{0: without, 1: with}{Knowledge: At the start of every minute}
    \bottomrule
    \end{tabular}
\end{table}

\subsection{Bird}

We constructed two datasets based on an open bird migration dataset \cite{Kranstauber2022}. According to the number of contextual data states, we named the two datasets as \texttt{Bird-2} and \texttt{Bird-8}, respectively.
Note that we set $|U|$ as 1 because the dataset did not record many birds' locations simultaneously.

\texttt{Bird-8} preserves eight contextual data states most relevant to birds' migratory trajectories.
For some data states, we were not sure whether it is reasonable to classify them into the current types (e.g., regard temperature as a ranged data state).
But this is supposed to have little influence on our model.

\texttt{Bird-2} only preserves two time-related contextual data states, namely, the timestamp and the season.

\begin{table}[H]
    \caption{All the data states considered as context data.}
    \label{tab:dds}
    \begin{tabular}{p{0.6in}p{1.6in}p{1.3in}p{3in}}
    \toprule
    Value Types & Data States & Normalization Params & Note \\
    \midrule

    \multirow{4}{*}{\vta}
    \dsr{surface temperature}{Min=250, MAX=320}{May be in Kelvin.}
    \cmidrule(lr){2-4}
    \dsr{air temperature}{Min=250, MAX=320}{The average air temperature is between 400m and 1000m.}
    \cmidrule(lr){2-4}
    \dsr{surface pressure}{MIN=97000, MAX=105000}{}
    \cmidrule(lr){2-4}
    \dsr{timestamp}{MIN=0, MAX=525600}{Considering that bird migration takes a year as a cycle, we convert the timestamp into the ordinal number of minutes in each year, obtaining a ranged data state from 0 to 365 * 24 * 60.}
    
    \cmidrule(lr){1-4}
    
    \multirow{3}{*}{\vtb}
    \dsr{wind speed in x direction}{$\hat{MAX}$=11}{}
    \cmidrule(lr){2-4}
    \dsr{wind speed in y direction}{$\hat{MAX}$=8}{}
    \cmidrule(lr){2-4}
    \dsr{bird density}{$\hat{MAX}$=6}{The vertically integrated bird density}
    
    \cmidrule(lr){1-4}
    
    \multirow{2}{*}{\vtc}
    \dsr{season}{0: Spring, 1: Autumn}{Birds usually migrate at these two seasons. The original dataset does not have data in the other two seasons.}
    
    \bottomrule
    \end{tabular}
\end{table}

%% file: content/a3_exp_results.tex
\pagebreak

\section{Complete Experiments Results}
\label{app:exp}

\subsection{Comparative Experiment I}

\begin{table}[h]
    \caption{The ADE loss of ten-round experiments on \texttt{DOTA}.}
    \tabcolsep=3pt
    \begin{tabular}{crrrrrrrrrrrr}
        \toprule
        Model & Round 0 & Round 1 & Round 2 & Round 3 & Round 4 & Round 5 & Round 6 & Round 7 & Round 8 & Round 9 & MEAN & SD \\
        \midrule
        \texttt{LSTM} & 1476.79 & 1393.91 & 1495.63 & 900.35 & 1185.02 & 1187.28 & 1048.58 & 1203.04 & 1555.00 & 1347.27 & 1279.29 & 199.89 \\

        \texttt{Trans} & 1080.47 & 1222.13 & 1316.80 & 1053.44 & 1027.95 & 1317.84 & 1224.36 & 1420.34 & 1243.66 & 1370.29 & 1227.73 & 128.74 \\
        
        \texttt{Tjf} & 1172.08 & 936.51 & 876.07 & 920.63 & 905.60 & 1061.73 & 1060.32 & 1030.19 & 861.18 & 869.66 & 969.40 & 99.73 \\
        
        \texttt{DATF} & 749.47 & 867.41 & 724.00 & 1011.85 & 591.48 & 798.94 & 721.36 & 838.30 & 851.53 & 959.48 & 811.38 & 116.48 \\
        
        \texttt{IETP-10} & 723.15 & 809.29 & 593.37 & 603.25 & 497.87 & 839.06 & 570.39 & 621.59 & 540.13 & 456.66 & 625.48 & 120.66 \\
        
        \texttt{IETP-20} & 479.89 & 524.78 & 615.08 & 489.99 & 605.00 & 574.85 & 583.03 & 579.90 & 502.46 & 614.51 & 556.95 & 49.95 \\

        \midrule
        
        \texttt{CATP-10} & 567.74 & 657.37 & 636.25 & 650.59 & 440.55 & 444.40 & 554.02 & 651.52 & 486.70 & 548.33 & 563.74 & 80.61 \\
        
        \texttt{CATP-20} & 442.69 & 476.61 & 434.66 & 364.47 & 385.90 & 408.36 & 433.83 & 399.40 & 451.29 & 463.97 & 426.12 & 33.87 \\
        
        \texttt{CATP-10-Tjf} & 525.01 & 569.00 & 623.32 & 479.20 & 525.51 & 635.43 & 553.09 & 428.19 & 521.44 & 521.29 & 538.15 & 58.61 \\
        
        \texttt{CATP-20-Tjf} & 410.35 & 414.58 & 442.17 & 453.31 & 401.12 & 415.52 & 495.63 & 333.98 & 449.04 & 417.91 & 423.36 & 36.94 \\
        \bottomrule
    \end{tabular}
\end{table}

\begin{table}[h]
    \caption{The ADE loss of ten-round experiments on \texttt{Bird-8}.}
    \tabcolsep=3pt
    \begin{tabular}{crrrrrrrrrrrr}
        \toprule
        Model & Round 0 & Round 1 & Round 2 & Round 3 & Round 4 & Round 5 & Round 6 & Round 7 & Round 8 & Round 9 & MEAN & SD \\
        \midrule
        \texttt{LSTM} & 23.99 & 19.33 & 17.21 & 23.65 & 19.08 & 19.13 & 19.26 & 25.81 & 25.85 & 18.93 & 21.22 & 3.06 \\

        \texttt{Trans} & 21.97 & 18.41 & 18.21 & 18.46 & 23.74 & 18.42 & 22.21 & 24.74 & 16.79 & 23.81 & 20.68 & 2.76 \\
        
        \texttt{Tjf} & 18.20 & 21.81 & 17.28 & 16.70 & 19.67 & 22.80 & 17.16 & 20.52 & 19.81 & 20.38 & 19.43 & 1.95 \\
        
        \texttt{DATF} & 15.98 & 13.99 & 16.03 & 16.86 & 16.28 & 15.08 & 17.30 & 15.26 & 15.97 & 16.25 & 15.90 & 0.89 \\
        
        \texttt{IETP-10} & 17.25 & 18.13 & 16.27 & 16.85 & 15.43 & 16.75 & 15.47 & 16.46 & 15.59 & 16.19 & 16.44 & 0.81 \\
        
        \texttt{IETP-20} & 15.91 & 13.14 & 13.09 & 13.78 & 14.56 & 12.65 & 14.46 & 12.41 & 12.68 & 15.09 & 13.78 & 1.12 \\

        \midrule
        
        \texttt{CATP-10} & 10.93 & 11.60 & 12.16 & 12.51 & 12.31 & 12.01 & 11.72 & 12.09 & 12.39 & 11.65 & 11.94 & 0.45 \\
        
        \texttt{CATP-20} & 17.20 & 15.93 & 19.52 & 19.22 & 17.20 & 16.19 & 14.76 & 16.21 & 16.59 & 17.97 & 17.08 & 1.41 \\
        
        \texttt{CATP-10-Tjf} & 11.67 & 12.92 & 12.22 & 9.68 & 12.34 & 12.04 & 12.09 & 12.92 & 12.95 & 10.59 & 11.94 & 1.01 \\
        
        \texttt{CATP-20-Tjf} & 20.49 & 17.48 & 20.31 & 18.84 & 19.28 & 20.28 & 14.81 & 16.44 & 18.65 & 16.95 & 18.35 & 1.80 \\
        \bottomrule
    \end{tabular}
\end{table}

\begin{table}[h]
    \caption{The ADE loss of ten-round experiments on \texttt{Bird-2}.}
    \tabcolsep=3pt
    \begin{tabular}{crrrrrrrrrrrr}
        \toprule
        Model & Round 0 & Round 1 & Round 2 & Round 3 & Round 4 & Round 5 & Round 6 & Round 7 & Round 8 & Round 9 & MEAN & SD \\
        \midrule
        \texttt{LSTM} & 24.49 & 24.28 & 28.95 & 16.71 & 21.21 & 17.37 & 21.89 & 28.86 & 19.81 & 30.55 & 23.41 & 4.63 \\

        \texttt{Trans} & 22.85 & 26.99 & 26.09 & 24.55 & 23.91 & 24.32 & 21.01 & 21.21 & 26.22 & 27.01 & 24.42 & 2.10 \\
        
        \texttt{DATF} & 17.91 & 15.75 & 14.95 & 16.27 & 16.11 & 17.36 & 17.58 & 17.13 & 16.64 & 16.69 & 16.64 & 0.86 \\
        
        \texttt{IETP-10} & 18.63 & 17.53 & 20.50 & 20.75 & 20.54 & 22.41 & 22.22 & 19.53 & 21.15 & 19.48 & 20.27 & 1.44 \\
        
        \texttt{IETP-20} & 17.56 & 17.92 & 16.68 & 17.95 & 14.84 & 16.40 & 18.04 & 19.80 & 17.50 & 21.05 & 17.77 & 1.64 \\

        \midrule
        
        \texttt{CATP-10} & 15.28 & 15.13 & 17.79 & 20.90 & 18.31 & 22.32 & 19.27 & 14.22 & 13.90 & 20.25 & 17.74 & 2.83 \\
        
        \texttt{CATP-20} & 20.52 & 27.77 & 27.57 & 24.69 & 22.28 & 21.59 & 22.67 & 19.93 & 21.91 & 24.21 & 23.31 & 2.58 \\
        
        \texttt{CATP-10-Tjf} & 18.21 & 16.31 & 18.64 & 21.20 & 18.02 & 16.34 & 11.53 & 17.77 & 18.64 & 15.92 & 17.26 & 2.40 \\
        
        \texttt{CATP-20-Tjf} & 21.55 & 15.93 & 21.66 & 18.01 & 22.48 & 25.40 & 22.03 & 15.79 & 25.48 & 23.85 & 21.22 & 3.36 \\

        \midrule

        \multicolumn{13}{l}{\textit{*Note: Since \texttt{Tjf} accepted only $TX$ as the input, it was unnecessary to repeat the experiments for \texttt{Tjf} on \texttt{Bird-2}.}} \\
        
        \bottomrule
    \end{tabular}
\end{table}

\subsection{Comparative Experiment II}

\begin{table}[H]
    \caption{The MSE loss of ten-round experiments on \texttt{SP}.}
    \tabcolsep=3pt
    \begin{tabular}{crrrrrrrrrrrr}
        \toprule
        Model & Round 0 & Round 1 & Round 2 & Round 3 & Round 4 & Round 5 & Round 6 & Round 7 & Round 8 & Round 9 & MEAN & SD \\
        \midrule
        
        \texttt{LSTM} & 5.43 & 5.91 & 4.97 & 5.28 & 5.72 & 6.15 & 6.56 & 4.98 & 5.40 & 6.46 & 5.69 & 0.54 \\

        \texttt{Trans} & 5.85 & 5.25 & 5.08 & 5.85 & 5.23 & 6.96 & 5.03 & 5.67 & 6.17 & 4.67 & 5.57 & 0.63 \\
        
        \texttt{Informer} & 4.48 & 5.77 & 4.52 & 3.97 & 4.33 & 5.35 & 6.62 & 4.77 & 4.11 & 4.17 & 4.81 & 0.81 \\
        
        \texttt{Autoformer} & 2.70 & 3.29 & 3.45 & 3.22 & 3.84 & 3.10 & 2.81 & 3.16 & 2.99 & 3.06 & 3.16 & 0.31 \\
        
        \texttt{FEDformer} & 3.12 & 2.81 & 3.92 & 3.36 & 2.97 & 2.99 & 2.76 & 3.15 & 3.34 & 3.13 & 3.16 & 0.32 \\
        
        \texttt{TimesNet} & 2.97 & 3.06 & 2.73 & 2.98 & 3.16 & 3.16 & 2.48 & 2.62 & 2.90 & 3.22 & 2.93 & 0.23 \\

        \midrule
        
        \texttt{MW-LSTM} & 5.27 & 4.02 & 4.96 & 4.42 & 5.03 & 5.13 & 4.84 & 5.25 & 4.58 & 5.13 & 4.86 & 0.39 \\
        
        \texttt{MW-Trans} & 4.52 & 4.70 & 4.70 & 5.89 & 3.85 & 4.42 & 5.49 & 4.86 & 4.17 & 4.81 & 4.74 & 0.56 \\
        
        \texttt{MW-TN} & 2.90 & 2.70 & 2.78 & 3.01 & 2.84 & 3.05 & 2.91 & 2.84 & 2.61 & 2.92 & 2.86 & 0.13 \\
        \bottomrule
    \end{tabular}
\end{table}

\begin{table}[H]
    \caption{The MSE loss of ten-round experiments on \texttt{ET}.}
    \tabcolsep=3pt
    \begin{tabular}{crrrrrrrrrrrr}
        \toprule
        Model & Round 0 & Round 1 & Round 2 & Round 3 & Round 4 & Round 5 & Round 6 & Round 7 & Round 8 & Round 9 & MEAN & SD \\
        \midrule
        
        \texttt{LSTM} & 1.25 & 1.44 & 1.59 & 1.19 & 1.43 & 1.36 & 1.46 & 1.32 & 1.52 & 1.38 & 1.40 & 0.12 \\

        \texttt{Trans} & 1.28 & 1.38 & 1.31 & 1.24 & 1.05 & 1.03 & 1.39 & 1.23 & 1.09 & 1.31 & 1.23 & 0.13 \\
        
        \texttt{Informer} & 1.26 & 1.03 & 0.95 & 0.81 & 1.11 & 1.07 & 1.05 & 0.98 & 1.16 & 1.01 & 1.04 & 0.12 \\
        
        \texttt{Autoformer} & 0.48 & 0.53 & 0.53 & 0.52 & 0.44 & 0.53 & 0.53 & 0.46 & 0.56 & 0.41 & 0.50 & 0.05 \\
        
        \texttt{FEDformer} & 0.48 & 0.49 & 0.46 & 0.48 & 0.42 & 0.53 & 0.52 & 0.41 & 0.39 & 0.42 & 0.46 & 0.05 \\
        
        \texttt{TimesNet} & 0.60 & 0.42 & 0.47 & 0.46 & 0.41 & 0.47 & 0.42 & 0.48 & 0.48 & 0.45 & 0.47 & 0.05 \\

        \midrule
        
        \texttt{MW-LSTM} & 0.86 & 1.01 & 0.89 & 0.82 & 0.89 & 1.09 & 1.05 & 1.11 & 1.10 & 1.06 & 0.99 & 0.10 \\
        
        \texttt{MW-Trans} & 1.09 & 0.97 & 1.09 & 0.89 & 0.94 & 1.11 & 1.16 & 0.76 & 0.99 & 1.04 & 1.00 & 0.11 \\
        
        \texttt{MW-TN} & 0.51 & 0.48 & 0.39 & 0.45 & 0.46 & 0.45 & 0.46 & 0.43 & 0.36 & 0.50 & 0.45 & 0.04 \\
        
        \bottomrule
    \end{tabular}
\end{table}